# Deep Concept Identification for Generative Design

Ryo Tsumoto[1], Kentaro Yaji[1], Yutaka Nomaguchi[1], Kikuo Fujita[1]

[1]Department of Mechanical Engineering, Osaka University, 1-1 Yamadaoka, Suita, Osaka 565-0871 Japan.

Corresponding author: tsumoto@syd.mech.eng.osaka-u.ac.jp (Ryo Tsumoto)

**Abstract**

A generative design based on topology optimization provides diverse alternatives as entities in a computational model with a high design degree. However, as the diversity of the generated alternatives increases, the cognitive burden on designers to select the most appropriate alternatives also increases. Whereas the concept identification approach, which finds various categories of entities, is an effective means to structure alternatives, evaluation of their similarities is challenging due to shape diversity. To address this challenge, this study proposes a concept identification framework for generative design using deep learning (DL) techniques. One of the key abilities of DL is the automatic learning of different representations of a specific task. Deep concept identification finds various categories that provide insights into the mapping relationships between geometric properties and structural performance through representation learning using DL. The proposed framework generates diverse alternatives using a generative design technique, clusters the alternatives into several categories using a DL technique, and arranges these categories for design practice using a classification model. This study demonstrates its fundamental capabilities by implementing variational deep embedding, a generative and clustering model based on the DL paradigm, and logistic regression as a classification model. A simplified design problem of a two-dimensional bridge structure is applied as a case study to validate the proposed framework. Although designers are required to determine the viewing aspect level by setting the number of concepts, this implementation presents the identified concepts and their relationships in the form of a decision tree based on a specified level.

## 1. Introduction

Generative design is a computational design method that provides alternatives that satisfy the given design requirements [1,2]. These alternatives are represented as entities within a computational model and stimulate the designer's creativity in the early design stage [3]. Because this method generates various alternatives by parametrically varying the design geometry, the freedom of representation of the geometry is an important factor in the diversity of the generated alternatives. Several types of representations exist such as representative dimensions under the given form [2], control parameters of the given spline curve [4,5], and material distribution within the given design domain [6,7]. Among these, representation by material distribution is the most suitable for exploring diverse alternatives owing to its high design degree.

Design optimization techniques are effective means of exploring promising alternatives in a computational model, and these techniques are typically used in the later design phase. Topology optimization [8] is a structural optimization technique that determines optimal material distribution within a given design domain. This technique has been expanded into various fields, including fluid-based problems [9] and heat transfer problems [10]. Topology optimization can be utilized as a generator for diverse alternatives including multiple configurations in generative design [6,11]. To enhance the diversity of alternatives, generative design frameworks are proposed by incorporating topology optimization with artificial intelligence techniques such as reinforcement learning [12], deep neural networks [13,14], and deep generative models [7,15]. However, as the diversity of the generated alternatives increases, the cognitive burden on designers to select the most appropriate alternatives also increases.

In contrast, the early design stage is commonly understood as the conceptual design stage where a design concept is generated by combining physical principles expected to achieve the required function [16]. To this end, methods have been proposed to generate and evaluate alternatives systematically [17,18]. Physical principles provide designers with insights into the mapping relationships between the design parameters and performance criteria, representing various plans of entities. Whereas these principles enable designers to effectively consider a wide range of alternatives, what is the physical principle is often vague. In another approach, these principles are identified inductively from various design solutions. This concept identification approach finds various categories of entities that share some essential features as design concepts [19]. These categories act as design concepts when providing insights into the mapping relationships between design parameters and performance criteria, representing various entities. Because generative design usually provides a large number of alternatives based on implicit principles, this approach is suitable for considering them effectively.

Some methods that identify categories of alternatives provided by generative design have been proposed to evaluate various alternatives efficiently [20,21]. These methods categorize alternatives based on the similarities calculated from qualitative and quantitative features using clustering algorithms. On the other hand, the identified categories should provide insights into the mapping relationships between geometric properties and structural performance to act as a design concept. The concept identification method proposed by Lanfermann and Schmitt [19] extracts these relationships by introducing a criterion for consistency across multiple viewing aspects for categorization. Because alternatives that have similar structural performances but different geometric properties exist, geometric properties are the focus of concept identification for diverse alternatives. However, in conventional clustering algorithms, traditional similarity measures, such as the Euclidian distance, usually cannot provide meaningful categories for high-dimensional data such as diverse alternatives that are represented by material distributions [22]. Conversion into different low-dimensional representations is an effective approach to clustering high-dimensional data. Although some representations are considered for generative design [23,24], finding such representations is usually challenging for designers based on their experience.

To address this challenge, this study proposes a concept identification framework for generative design using deep learning (DL) techniques. One of the key abilities of DL is to automatically learn different representations of a specific task [25]. This study defines deep concept identification as finding various categories that provide insights into the mapping relationships between geometric properties and structural performance through representation learning using DL. The proposed deep concept identification framework generates diverse alternatives by a generative design technique, clusters alternatives into several categories using a DL technique, and arranges these categories for design practice using a classification model. This paper demonstrates its fundamental capabilities by implementing variational deep embedding (VaDE) [26], a generative and clustering model based on the DL paradigm, along with logistic regression as a classification model. A simplified design problem of a two-dimensional bridge structure is applied as a case study to validate the proposed framework.

The rest of this paper is structured as follows: Section 2 presents theoretical backgrounds by referring to related works and formalizes the deep concept identification framework. Section 3 implements the proposed framework using VaDE and logistic regression. Section 4 provides the experimental application to a simplified design problem of bridge structure. Section 5 discusses the capabilities of the proposed framework. Section 6 presents concluding remarks and future works.

## 2. Theoretical background
### 2.1 Concept identification in engineering design

What we refer to as "concepts" play an important role in considering a wide range of alternatives in conceptual design. However, there is no common understanding of what a concept is. The meaning of the concept is relatively straightforward from the standpoint of linguistics. Recognizing a vast number of objects in the real world at any time is impossible, each with a unique index. For effective recognition, they are categorized into classes that share some representative features, and the members of each category are described, that is, they are labeled with a unique index. Whereas the representatives of each category provide overviews of the objects, each object belonging to a certain category can be identified by adding secondary features to representative ones. Thus, those categories act as concepts to effectively recognize a vast number of objects. Based on this understanding, key concept identification from a set of documents is performed through semantic categorization [27,28]. In these cases, documents correspond to objects, and categories of documents correspond to concepts. Identified key concepts and their relationships to navigate effective information retrieval from a set of documents. In the engineering design process, entities correspond to objects, and categories of entities act as design concepts [29]. They and their relationships guide succeeding design processes as design knowledge

[30].

Because the results of categorization depend on the features focused on, various concepts can be identified to provide different insights. In engineering design, two types of features are typically considered [31]. One is the attributes corresponding to the features for specifying an entity, such as its shape or geometry. The other is the behavior corresponding to the features for evaluating an entity, such as engineering or aesthetic performance. Because design requirements are described in terms of behaviors, whereas design solutions are described in terms of attributes, the mapping relationships between them act as design knowledge for selecting the most appropriate design solution [32]. Consequently, conceptual design can be defined as the activity of finding concepts related to attributes that are expected to enable the requirement represented by concepts related to behavior. However, the relationship between the concepts of both sides cannot be understood without linking them to corresponding entities, because they are conceptualized on independent viewing aspects. Therefore, concept identification in engineering design is an entity-level task that categorizes a set of entities while mapping their attributes and behaviors.

General design theory (GDT) [33] can be used to formalize the aforementioned understanding of concept identification. The GDT assumes that a concept categorizes a set of entities into two subsets that are true and false under a certain condition. Furthermore, each entity is identified as a set of concepts. Based on GDT, a design activity can be defined as finding any entity as a design solution in the product of classifications related to the behavior that represents a given design requirement. If multiple entities exist in the product of classifications related to behavior, a classification must be introduced to squeeze these entities into a unique one. Here, categories of entities are focused on rather than unique ones in conceptual design. These categories should share some essential attributes to ensure consistency in succeeding design processes. While designers intuitively perform conversions between concepts and entities during this process, the mutual relationship between them is an important issue in concept identification. Designers structure the plans of entities to evaluate their similarities or differences and implicitly form concepts as products of classification. The existence of entities triggers the described process of concept identification. As the results of concept identification depend on the scope of the entities, collecting the entities comprehensively is essential.

Therefore, the concept identification framework for conceptual design should comprise the following points: (1) Collecting various entities related to a design problem, (2) Categorizing collected entities based on their attributes while structuring to evaluate their similarities or differences, and (3) Arranging collected entities as the product of classifications related to their behavior.

## 2.2 Formalizing deep concept identification for generative design

Following the previous subsection, this study proposes deep concept identification as a framework for mapping concepts and entities in generative design. An overview of deep concept identification is shown in Fig. 1.

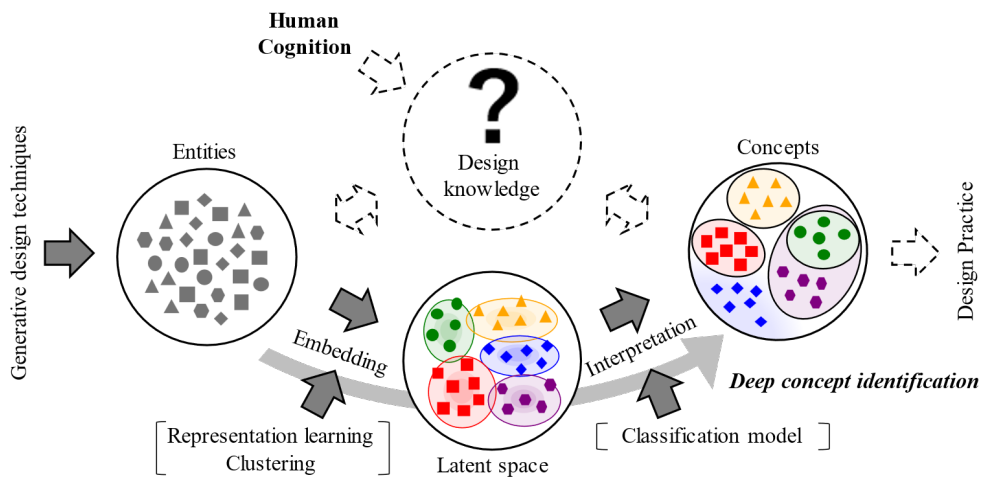

Fig. 1 Overview of deep concept identification

A certain number of entities are present in the circle on the left of Fig. 1, and concepts related to these entities are represented by the product of the four classifications in the circle on the right of Fig. 1. Although the mapping relationship between concepts and entities is acquired as design knowledge through human cognition, as shown in the upper half of Fig. 1, the proposed framework identifies this relationship using the procedure shown in the lower half of Fig. 1. Generative design techniques provide various alternatives as entities that can trigger deep concept identification. Methods based on topology optimization that explore the material distribution within a given design domain are effective for collecting a wide range of entities. Each collected entity is a specified material distribution that is usually represented in pixel or voxel format.

To arrange as the products of classifications, which is the goal of deep concept identification, the collected entities are categorized while being structured to evaluate their similarities or differences. First, two types of common computational approaches for categorization exist. One is clustering, which is an unsupervised approach for grouping data such that data in the same cluster are similar and data in different clusters are different to the maximum extent. This approach is generally used to structure given data [34]. The second is classification, which is a supervised approach for training a classifier corresponding to predefined groups. This approach is generally used to analyze the features of predefined groups and predict a group of new data [35]. Because no prior knowledge to guide categorization exists, clustering is a suitable means of categorizing the collected entities. Second, the data representation is an important factor in the performance of computational approaches. Because the attributes of the collected entities are represented by a high-dimensional representation such as a pixel or voxel format, similarities or differences cannot be effectively evaluated due to sparsity [36]. In such cases, data are typically converted into different low-dimensional representations. Representation learning is an approach used to discover a latent representation for a task using machine learning (ML) [37]. This approach is a fundamental concept of DL techniques that converts data into various representations through multilayer neural networks [25]. In the proposed framework, the collected entities are categorized based on their attributes and embedded into the latent space using representation learning and clustering.

After categorizing the collected entities based on their attributes, they are arranged as a product of behavior-related classifications. The categorization results guide the computational classification of entities for supervision. A critical issue in behavior-based classification is whether a unique category can be squeezed. However, defining a sufficient set of criteria without prior knowledge is challenging. The latent representations discovered by representation learning are features that effectively represent the collected entities. A new criterion can be identified by interpreting the discovered latent representation.

## 3. Fundamental implementation of deep concept identification

This study verifies the fundamental capabilities of deep concept identification by implementing the framework shown in Fig. 2.

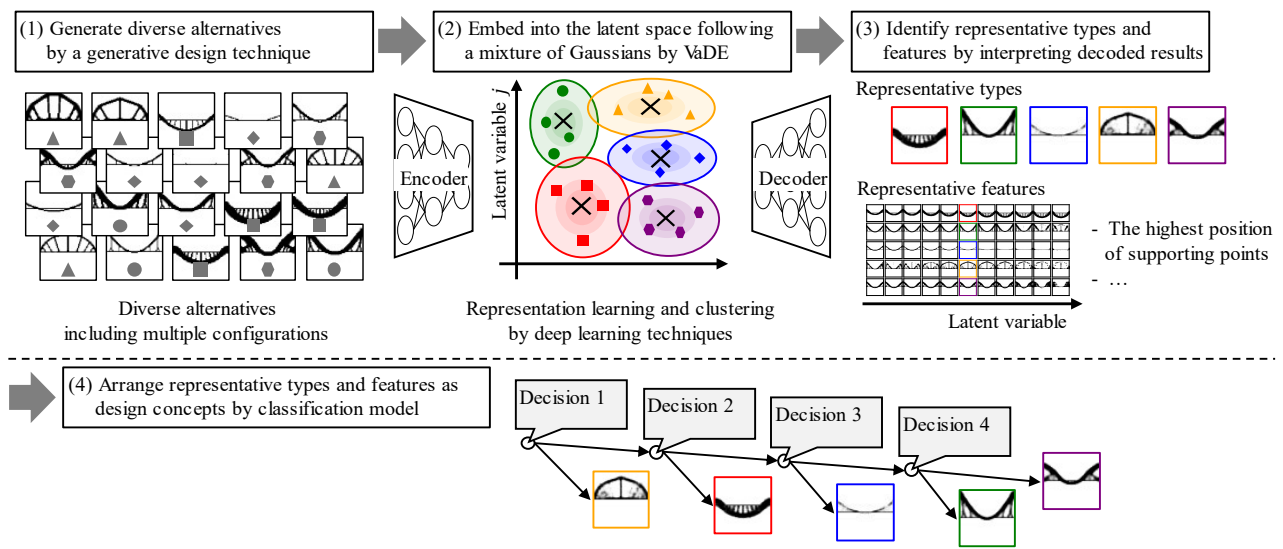

Fig. 2   Overview of the implementation of deep concept identification

The proposed framework comprises the following four steps: (1) Generate diverse alternatives using a generative design technique, (2) Embed them into the latent space following a mixture of Gaussians by VaDE, (3) Identify representative types and features by interpreting decoded results, and (4) Arrange representative types and features as design concepts using the classification model.

A mathematical representation is introduced following the design optimization paradigm to explain the details of the implementation. An optimal design problem is defined as finding design variables $x$ that minimize a set of objective functions $f_i(x)$ subject to a set of constraints $g_j(x)$. Both the objective functions and constraints are related to structural performance. Thus, this study calls the functions used as objective functions and constraints evaluation criteria $f_i(x)$ $(i = 1, ..., I)$ collectively. In this form, the attribute of $n$-th alternative is represented by $x_n$ and the behavior of $n$-th alternative is represented by $f_i(x_n)$.

Step (1) corresponds to the circle on the left in Fig. 1. In this step, generative design techniques based on topology optimization provide a set of alternatives $x_n$. A design problem is formulated under a topology optimization paradigm. First, the design domain, which is the area where materials can be distributed, is defined. Here, design variable $x$ is the material distribution in the defined design domain. One of the approaches to generating diverse alternatives is to solve a multi-objective design problem under various conditions. When a design problem is formulated as a multi-objective problem, a set of compromised solutions is provided under conflicts between objectives. Complicated cause-and-effect relationships between the material distribution and structural performance guide diverse alternatives that include multiple configurations in this approach.

Step (2) corresponds to the transition from the left-hand side circle to the lower-side circle in Fig. 1. In this step, VaDE models are trained to structure the generated alternatives through representation learning and clustering. For clustering based on attributes, design variables $x_n$ are used as the data representation of each alternative $n$. An autoencoder is a typical representation learning model. An autoencoder model comprises two types of neural networks, encoder and decoder. An encoder converts the data into a different representation, whereas a decoder converts the data into an original representation. The autoencoder model is trained to reconstruct the original data by the decoder from the new representation transformed by the encoder. An autoencoder can learn a specific latent representation by introducing a probabilistic model of data distribution in the latent space into the generative process by the decoder. These models are known as variational autoencoders (VAEs). The original VAE learns statistically independent latent representations that follow an isotropic unit Gaussian distribution as prior [38]. VaDE is a generalized model of VAE for clustering tasks [26]. Unlike VAE, VaDE learns statistically independent latent representations that follow a mixture-of-Gaussian distribution, instead of a single Gaussian distribution. This probabilistic model encourages similar data to form clusters in the latent space. The trained VaDE model clusters samples based on the posterior of each Gaussian. Therefore, VaDE can categorize generated data by simultaneously performing representation learning and clustering.

Step (3) corresponds to the preparation of the transition from the lower-side circle to the left-hand side circle in Fig. 1. To arrange the categories of alternatives as the product of classifications related to their behavior, the representative types and features of these categories are identified. A key advantage of VAEs is their ability to generate new data using a trained decoder while parametrically manipulating the latent representation. Because VaDE models the generative process in the same way as VAEs, the representative data of each cluster can be generated by decoding from the means of each Gaussian. These representatives show the attributes of each category of alternatives. Furthermore, VAEs can discover interpretable latent factors by learning a disentangled latent representation, depending on the nature of the assumed prior distribution of the latent representation [39,40]. VaDE is expected to discover interpretable latent factors owing to the nature of the independent prior distribution of the latent representations. These latent factors can be used to effectively distinguish between categories of alternatives. New criteria related to the behavior of alternatives are defined based on these latent factors and added a set of evaluation criteria $\{f_i(x)\}$.

Step (4) corresponds to the transition from the lower-side circle to the left-hand side circle shown in Fig. 1. To identify concepts as products of classification, the generated alternatives are arranged based on the results of categorization in the previous step by the classification model. The evaluation criteria $\{f_i(x_n)\}$ are used as data representation, and clustering results by VaDE are used as a label for each alternative $n$. The interpretability of the classification model is a crucial factor in arranging the identified concepts as design knowledge. However, conflicts exist between the performance and interpretability of the classification model [41]. For instance, although a classification model constructed with neural networks can achieve high classification performance owing to the numerous trainable parameters and non-linear transformations within each network, interpreting these trained parameters and transformations is impossible. In contrast, a linear classification model has limited classification performance but high interpretability owing to its simplicity.

## 4. Experimental application
### 4.1 Definition of a simplified conceptual design problem of bridge structure

To demonstrate the capabilities of deep concept identification, the proposed framework is applied to a simplified design problem of bridge structures. A bridge is a structure that supports a roadway over obstacles using piers [42]. Bridge structures are categorized into some types such as beams, trusses, arches, and suspensions based on how they support the roadway. These types are considered design concepts because they are a consequence of the categorization of various existing bridge designs. Therefore, this study considers that the conceptual design of a bridge structure corresponds to the selection of the most appropriate type as the design concept. The most appropriate type is selected based on the structural performance and aesthetic criteria under specific conditions such as load and location. Because these types are mostly characterized by a material distribution over the cross-sectional area between the two piers, the evaluation criteria can be considered based on the material distribution over this area in the conceptual design.

This application focuses on two-dimensional design problems of material distribution over the cross-sectional area between two piers as a conceptual design problem of bridge structures. The roadway is supported by the left and right sides of the design domain corresponding to the piers. For simplicity, while the loads on the roadway are generally variable or multi-modal, they are assumed to be uniformly distributed. Furthermore, although design domains with various aspect ratios are possible, only the square design domain is considered. The material distribution is assumed to be symmetrical to the vertical line through the center of the design domain.

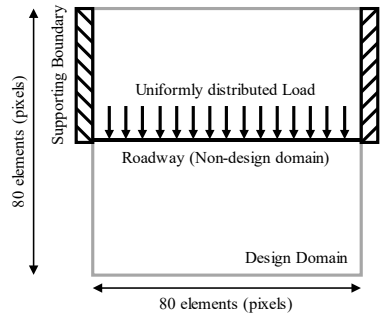

Fig. 3   Design domain for the conceptual design problem of bridge structure

The design domain of the simplified conceptual design problem of a bridge structure is shown in Fig. 3. The bridge structure is represented by a material distribution within the square design domain, the height and width of which are 80 elements (pixels) for implementation of topology optimization, which is explained in the next section. A horizontal roadway passed through the center of the design domain as a non-design domain. The height is 1 element (pixel). The roadway can be supported by highlighted boundaries in the upper halves of the left and right sides of the design domain. The selected boundaries for support are provided as conditions. An alternative is to be optimized for the stiffness and material volume under a support condition while manipulating the material distribution.

### 4.2 Generating diverse alternatives of bridge structures by generative design

This application uses density-based topology optimization to find optimal material distributions. The basic idea is to introduce continuous design variables for optimization using a gradient-based method. Each $i$-th design variable $x_i$ represents the density of the $i$-th finite element assigned to the design domain, and takes a value from 0 to 1. A value of 1 corresponds to the material, whereas a value of 0 corresponds to a void, and an intermediate value does not indicate manufacturability. To calculate the stiffness of the material distribution in the design domain using such design variables, the Young's modulus of each finite element is interpolated as shown in the following Eq. (1):

$$E_i(x_i) = E_{\min} + x_i^p(E_0 - E_{\min}) \tag{1}$$

where $E_0$ denotes the material stiffness and $E_{\min}$ denotes a very small stiffness introduced to prevent the stiffness matrix from becoming singular when the design variable becomes zero. $p$ denotes a penalization factor that avoids intermediate density values.

Whereas the defined design problem is a multi-objective problem that optimizes the stiffness and material volume, the material volume is considered a constraint to be satisfied in a gradient-based method. A set of compromised solutions are obtained by optimizing the stiffness while varying the volume fraction of the volume constraint. The topology optimization problem of the bridge structure is formulated as follows:

$$\text{Find} \quad \mathbf{x} = (x_1, \dots, x_{n_d})^{\mathrm{T}}, \tag{2}$$

$$\text{that minimizes} \quad f(\mathbf{x}) = \mathbf{U}^{\mathrm{T}}\mathbf{K}\mathbf{U} = \sum_{i=1}^{n_d} E_i(x_i)\mathbf{u}_i^{\mathrm{T}}\mathbf{k}_0\mathbf{u}_i, \tag{3}$$

$$\text{subject to} \quad g(\mathbf{x}) = \sum_{i=1}^{n_d} x_i s_i \leq \bar{g}, \tag{4}$$

$$\mathbf{K}\mathbf{U} = \mathbf{F}, \tag{5}$$

$$0 \leq x_i \leq 1 \quad \text{for } i = 1, \dots, n_d, \tag{6}$$

where $x_i$ denotes the $i$-th component of the design variable vector $\mathbf{x}$ and represents the density of $i$-th finite element, $s_i$ denotes the volume of the $i$-th finite element, and $n_d$ denotes the number of finite elements and dimensions of the design variables. $\mathbf{U}$ and $\mathbf{F}$ denote the global displacement and force vectors, respectively, $\mathbf{K}$ denotes the global stiffness matrix, $\mathbf{u}_i$ denotes the displacement vector of the $i$-th finite element, and $\mathbf{k}_0$ denotes the element stiffness matrix. $f(\mathbf{x})$ denotes the mean compliance, which is a criterion corresponding to the stiffness and $g(\mathbf{x})$ denotes the material volume. $\bar{g}$ denotes the allowable volume fraction, and Eq. (4) corresponds to the volume constraints. A support condition is imposed as the Dirichlet boundary condition in Eq. (5).

The code for the density-based topology optimization is implemented based on the 88 lines of MATLAB code [43]. A finite element method with 80 × 80 square elements is used for the evaluation. The initial design variables are uniformly set to satisfy the volume constraints. The design variables are updated using the optimality criteria method with the move limit of 0.2. The penalization factor is set to 3, and sensitivity filtering with a filter radius of 1.5 is used.

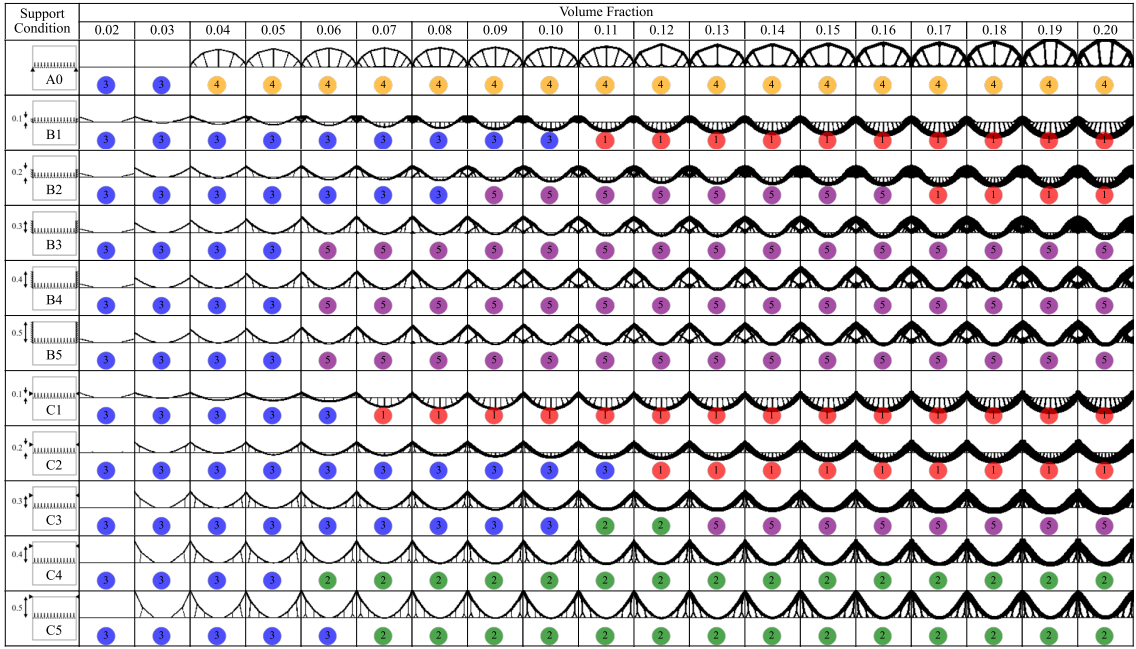

Fig. 4  Generated alternatives by topology optimization under the various support conditions and volume fractions

This application generates various alternatives by varying the volume fraction and the support conditions as design parameters. The results of the generation of alternatives are shown in Fig. 4. The leftmost column of Fig. 4 shows the support conditions, the number of which is 11. The design domain for support condition A0 is exceptionally limited to the upper half because of its vertically symmetric property. The topmost row of Fig. 4 shows the volume fractions of the volume constraints, the number of which is 19 from 0.02 to 0.2 incremented by 0.01. The cells in Fig. 4 show the 209 alternatives. The colored circles in each cell represent the clustering results obtained using VaDE. To consider manufacturability in the succeeding processes, each alternative is binarized to eliminate intermediate values using the Otsu method [44], and the mean compliance under support condition B5 and the volume are recalculated.

## 4.3 Embedding alternatives into latent space and identifying representative types and features by the training VaDE model

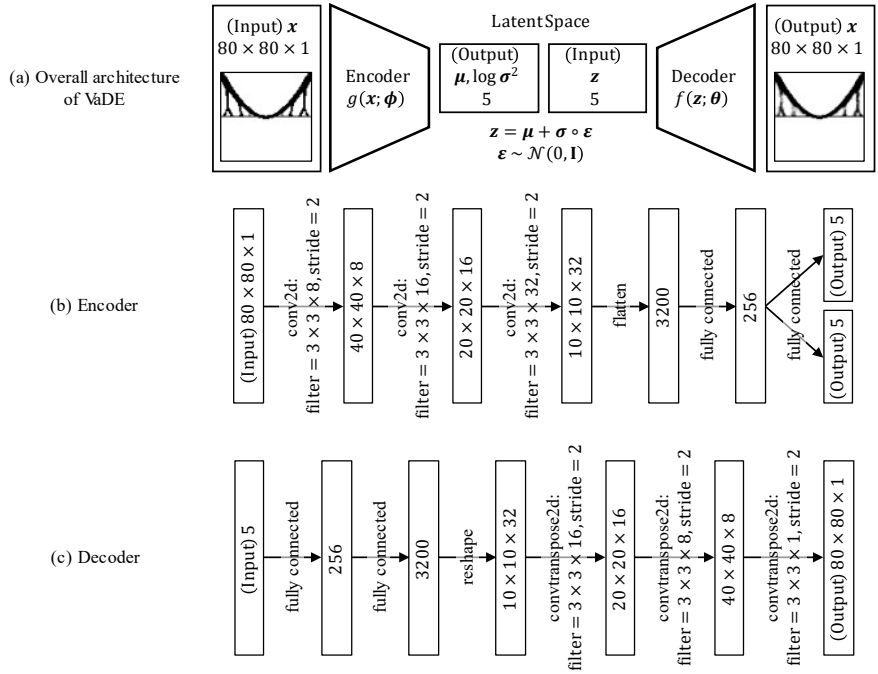

Fig. 5  Architecture of VaDE model

A VaDE model is trained to categorize generated 209 alternatives based on the design variables. The architecture of the VaDE model is illustrated in Fig. 5. Because each alternative is represented as an 80 × 80 gray-scale bitmap image by mapping each square element to a pixel, the input and output of VaDE are set to 80 × 80 × 1. An encoder with three convolutional layers and two fully connected layers compresses the images into low-dimensional latent variables. The decoder has a symmetrical structure to the encoder by replacing the convolutional layers with transposed convolutional layers.

Following the original study on VaDE, this application first trains the VAE model with the same architecture as that shown in Fig. 5 to improve the reconstruction ability. The VAE model is trained using the following loss function:

$$L_{VAE} = -D_{KL}\left(q_\phi(z|x)||p_\theta(z)\right) + \log p_\theta(x|z) = L_{KL} + L_{rec} \qquad (7)$$

where $x$ denotes the design variable, $z$ denotes the latent variable, and $\theta$ and $\phi$ denote the parameters of encoder and decoder, respectively. The first term $L_{KL}$ denotes the Kullback–Leibler(KL) divergence that corresponds to the criterion used to evaluate the similarity between the distribution of data converted by the encoder in the latent space and a prior distribution, that is, an isotropic unit Gaussian distribution. The second term $L_{rec}$ corresponds to the reconstruction loss of the encoder and decoder. To select the dimensions of the latent variables, this application compares the reconstruction loss of VAE models with 2-, 3-, 4-, 5-, 6-, 7-, 10-, 15-, and 20-dimensional latent variables. An Adam optimizer with a mini-batch size of 32 is used to update the parameters of the encoder and decoder. The number of epochs for VAE training is set to 150. The results of the VAE training are shown in Fig. 6. Although the reconstruction loss is improved by increasing the number of dimensions of the latent variables to five, no significant improvement is observed beyond five. To avoid the duplication of each discovered factor related to the latent variables, this application sets the latent dimension to five.

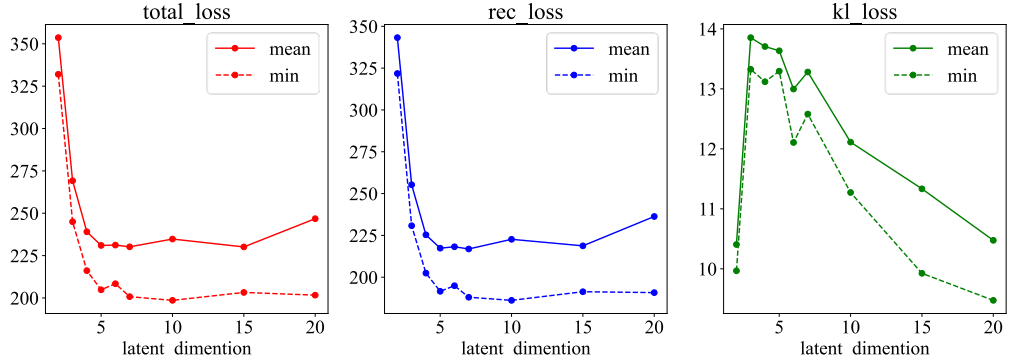

Fig. 6    Reconstruction loss of VAE

This application trains the VaDE model using the architecture shown in Fig. 5. The loss function of VaDE is similar to that of VAE, however, the first term $L_{KL}$ in Eq. (7) is replaced with the KL divergence to evaluate the similarity between the distribution of data converted by the encoder in the latent space and a mixture of Gaussian distributions. The number of clusters is set into 3, 5, and 7. An Adam optimizer with a mini-batch size of 32 is used to update the parameters of the encoder and decoder. The number of epochs for VaDE training is set to 300. Statically representative data are generated from the center of each cluster in the latent space using the trained VaDE model, as shown in Fig. 7. For the experimental discussion, this application sets the number of clusters to five in the subsequent processes. The clustering results of the alternatives with five clusters are indicated by a colored circle in Fig. 5.

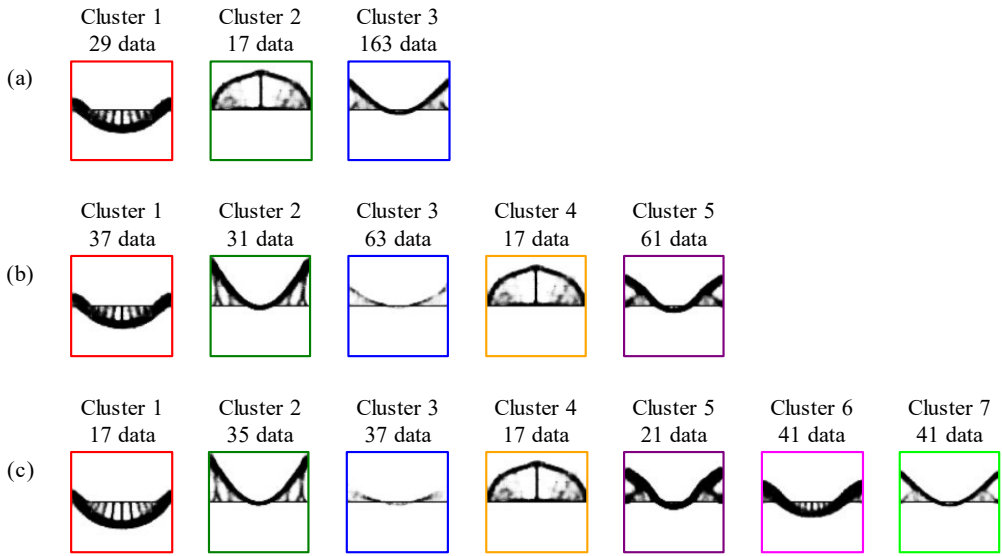

Fig. 7    Representative types when the numbers of clusters are set to (a) 3, (b) 5, and (c) 7.

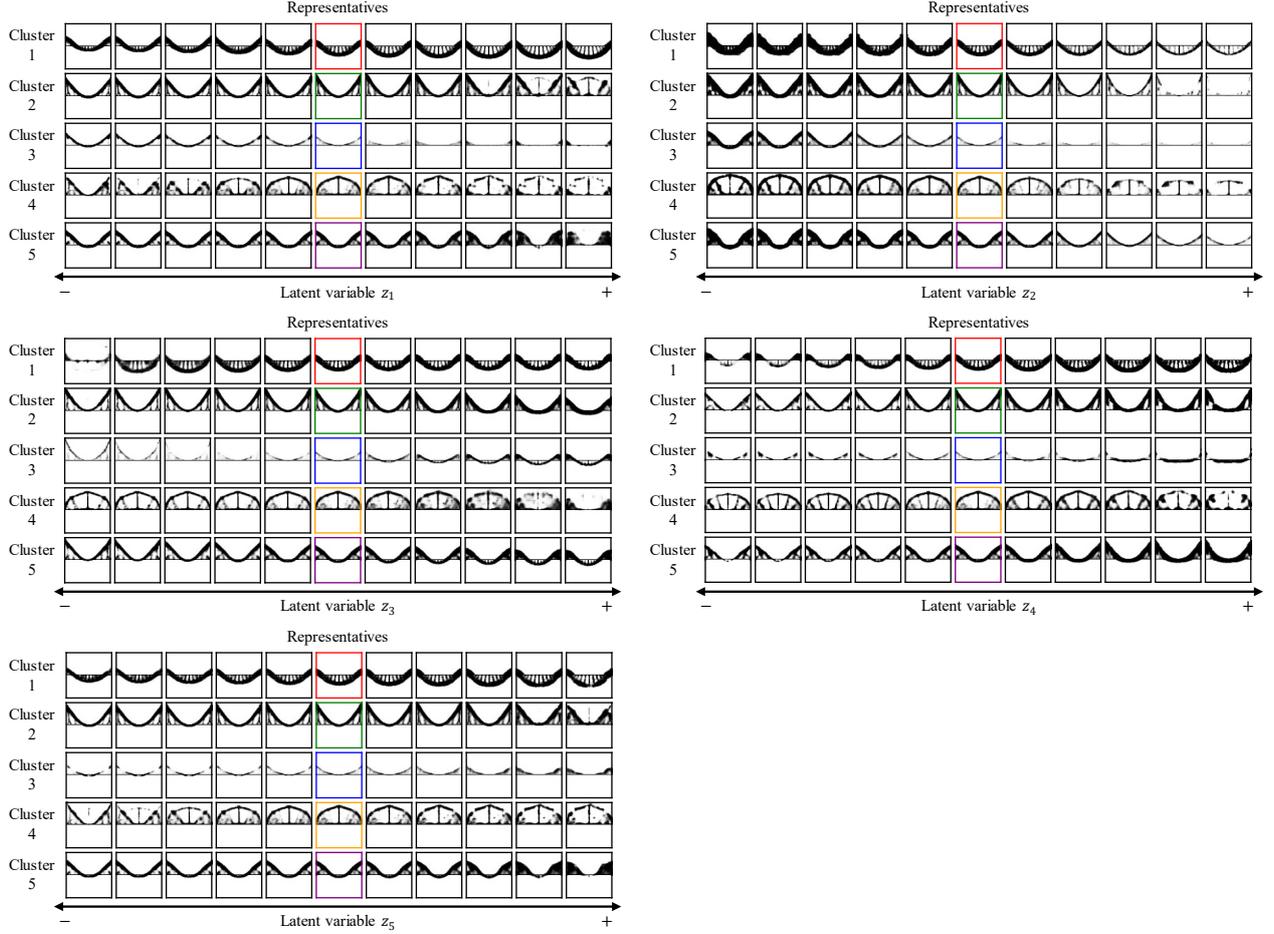

Fig. 8  Result of visualization by decoding while manipulating each latent variable for five representative configurations

To interpret the latent factors discovered from the latent variables toward the definition of the evaluation criteria, changes in the design variables are visualized by decoding while manipulating each latent variable independently. The result of visualization by decoding while manipulating each latent variable from -10 to 10 independently for five representative configurations is shown in Fig. 8. For latent variable $z_1$, the highest position of the supporting points gradually decreased as $z_1$ increased. For latent variable $z_2$, the volume fraction gradually decreased $z_2$ increased. For latent variable $z_3$, the centroid of the material distribution gradually decreased $z_3$ increased. However, for latent variables $z_4$ and $z_5$, no tendency related to specific structural features was observed. Consequently, the highest positions of the supporting points and the centroid of the material distribution were discovered as latent factors.

**4.4 Arranging representatives as design concepts by the classification model**

To arrange alternatives as the product of classifications related to behavior, the evaluation criteria for their behaviors are defined as follows, based on the formulation of topology optimization and discovered latent factors.

$f_1$: Mean compliance, which is the objective function of the topology optimization defined by Eq. (3), and corresponds to the criterion for stiffness.
$f_2$: Material volume, which is the constraint of topology optimization defined by Eq. (4), and corresponds to a criterion for cost efficiency.
$f_3$: Height of the centroid of material distribution, which is calculated from design variables, and may correspond to aesthetics.
$f_4$: Heights of the supporting points, which are calculated from the design variables correspond to the left-most elements of the design domain. This may correspond to ease of construction.

For statistical analyses, the values of each evaluation criterion are normalized such that the average becomes 0 and the standard deviation becomes 1.

This application arranges representative types and features using binary linear classifiers for interpretability. For this purpose, a set of alternatives labeled based on the clustering results are classified based on their evaluation criteria. A set of alternatives is divided into two subsets, such that alternatives included in the same cluster are included in the same subset. Class label $y$ of the $n$-th alternative is defined as follows:

$$y_n = \begin{cases} 1 & \text{if } n\text{-th alternative is included in the first subset} \\ 0 & \text{if } n\text{-th alternative is included in the second subset} \end{cases} \tag{8}$$

Under this class label, a binary linear classifier with four evaluation criteria is introduced as in the following Eq. (9):

$$s = \boldsymbol{w}^T \boldsymbol{f} = w_0 + w_1 f_1 + w_2 f_2 + w_3 f_3 + w_4 f_4 = \begin{cases} \geq 0 & \text{if } y = 1 \\ < 0 & \text{if } y = 0 \end{cases} \tag{9}$$

where $w_i$ $(i = 0, \ldots, 4)$ are the $i$-th components of coefficient $\boldsymbol{w}$ for identifying it. $f_i$ $(i = 0, \ldots, 4)$ is the $i$-th component of the evaluation criterion $\boldsymbol{f}$, with $f_0$ set to 1. The best possible binary classification case is somehow selected. Subsequently, if one subset corresponds to multiple categories, the alternatives of such subsets are classified in similarly. By repeating the binary linear classification until one subset corresponds to one category, all concepts are identified as the products of the classifications.

Logistic regression [45] is used to systematically identify the most appropriate coefficients. The logistic regression model introduces the following function known as the logistic sigmoid function to represent the posterior for $y = 1$:

$$p(y = 1 | \boldsymbol{f}) = \sigma(\boldsymbol{w}^T \boldsymbol{f}) = \frac{1}{1 + \exp(-\boldsymbol{w}^T \boldsymbol{f})} \tag{10}$$

This model determines the coefficients in Eq. (9) by maximizing the log-likelihood $\ln p(y|\boldsymbol{w})$. Although accuracy, calculated as the rate of correctly predicted data, is a well-known metric for evaluating classification models, this log-likelihood is suitable for evaluation because of the inherent learning process of logistic regression. The best case in possible binary classifications is selected systematically using the following steps: (1) Training logistic regression models for all possible cases by dividing the considered alternatives into two subsets according to Eq. (8). (2) Selecting the best case based on the log-likelihood of the trained logistic regression models. (3) Classifying the considered alternatives into two subsets using the selected logistic regression model.

The binary linear classification models are interpreted based on their coefficients. For instance, whereas the absolute value of the coefficient $|w_i|$ indicates the contribution of a related evaluation criterion $f_i$ to a classification, its sign indicates the direction of the contribution.

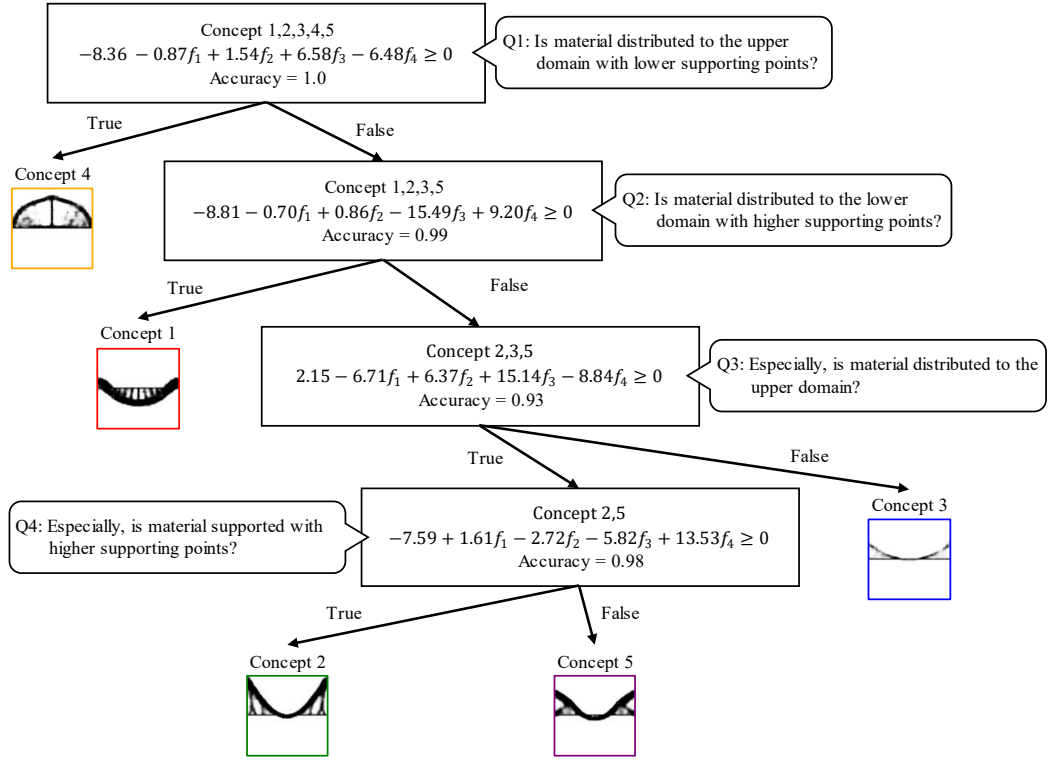

Fig. 9 Concepts of generated alternatives as the product of classifications

The result of the best case classification based on the evaluation criteria is shown in Fig. 9. Each concept corresponds to a cluster provided by VaDE. The four rectangles represent the classification models. Each classification model and concept are connected according to the classification paths. The first line in each rectangle shows the concepts of the considered alternatives, the second line shows the binary linear classifier trained using logistic regression, and the third line shows the accuracy of the classification model. Additionally, the interpretation results are shown next to each classification model for speech bubbles.

First, when applying this method to Concepts 1, 2, 3, 4, and 5, the following classifier that divides them into Concept 4 and the others is identified to be the most reasonable:

$$-8.36 - 0.87f_1 + 1.54f_2 + 6.58f_3 - 6.48f_4 = 0 \qquad (10)$$

Second, when applying this method to Concepts 1, 2, 3, and 5, the following classifier that divides them into Concept 1 and the others is identified as the most reasonable:

$$-8.81 - 0.70f_1 + 0.86f_2 - 15.49f_3 + 9.20f_4 = 0 \qquad (11)$$

Third, when applying this method to Concepts 2, 3, and 5, the following classifier that divides them into Concepts 2 and 5 and Concept 3 is identified as the most reasonable:

$$2.15 - 6.71f_1 + 6.37f_2 + 15.14f_3 - 8.84f_4 = 0 \qquad (12)$$

Finally, when applying this method to Concepts 2 and 5, the following classifier that divides them into Concept 2 and Concept 3 is identified as the most reasonable:

$$-7.59 + 1.61f_1 - 2.72f_2 - 5.82f_3 + 13.53f_4 = 0 \qquad (13)$$

The classification models are interpreted based on the signs and absolute values of their coefficients. The classifier defined in Eq. (10) identifies Concept 4 as characterized by a material distribution in the upper domain with lower supporting points compared to Concepts 1, 2, 3, and 5. The classifier defined in Eq. (11) identifies Concept 1 as characterized by material distribution in the lower domain with higher supporting points compared to Concepts 2, 3, and 5. The classifier defined in Eq. (12) identifies Concepts 2 and 5 as characterized by a material distribution that is distributed in the upper domain compared with Concept 3. The classifier defined in Eq. (13) identifies Concept 2 as characterized by a material distribution supported by higher supporting points than in Concept 5. The interpreted decision tree shown in Fig. 9 is expected to provide valuable knowledge for the conceptual design [32]. For instance, in cases where the design problem focuses on visual stability, the material distribution in the lower domain is appropriate. Designers can systematically identify Concept 1 as the most appropriate concept by selecting 'False' for the first question and 'True' for the second question.

## 5. Discussion

While only 209 alternatives are generated in the experimental application, as shown in Fig. 4, the deep concept identification is natively intended to be applied to a large number of alternatives. The experimental application was intended to focus on investigating whether concepts could be identified from diverse alternatives, including multiple configurations. In the case of simplified bridge structures, because the number of alternatives is not very large, representative configurations and their features can be identified based on visual analysis. This situation allows for the evaluation of the proposed framework. In contrast, VaDE and logistic regression, which are utilized for implementing the proposed framework, are based on the ML paradigm, which is superior in recognizing patterns from large amounts of data. Thus, although the computational cost increases, the proposed framework can be applied to a large number of alternatives.

The generated alternatives should be categorized based on their attributes while structuring them to evaluate their similarities or differences in concept identification. Deep concept identification embeds alternatives into a latent space through representation learning and clustering for categorization. As shown by the experiments on the reconstruction ability in Fig. 6, the appropriate number of dimensions of the latent space can generally be four or more. Thus, although the structure of alternatives in a latent space is an important factor in interpreting the results of categorization, it cannot be directly visualized. However, evaluating the clustering result based on whether the attributes of the alternatives in the same cluster are similar is relatively straightforward, as shown in Fig. 4. Therefore, performing representation learning and clustering simultaneously to evaluate the structurization of alternatives is essential. The VaDE is suitable for this purpose.

Another challenge in categorizing the generated alternatives is determining the appropriate number of identified concepts. Because the classification result depends on the viewing aspect level, design concepts introduce hierarchical structures into alternatives. The appropriate level is determined by following the design process phase. The viewing level corresponds to the number of clusters in the VaDE models. Whereas VaDE cannot learn hierarchical structures into alternatives during training, designers can explore the appropriate level iteratively by training VaDE while varying the number of clusters, as shown in Fig. 7. Detailed categories are obtained by increasing the number of clusters. For instance, when the number of clusters is increased from three to five, Cluster 1 with three clusters corresponds to Cluster 1 with five clusters, Cluster 2 with three clusters corresponds to Cluster 4 with five clusters, and Cluster three with 3 clusters is subdivided into Cluster 2, 3, and 5 with five clusters.

The latent factors discovered by DL lead to the definition of evaluation criteria for arranging the categories of entities as concepts in deep concept identification. In this experimental application, the latent factors discovered by VaDE are interpreted by visualizing the changes in the attributes while manipulating each latent variable independently, as shown in Fig. 8. The overall trends of the alternatives are effectively displayed by visualizing the representative types. However, whether latent factors can be interpreted from changes in attributes and linked to behavior to define evaluation criteria depends on the interpretability of the latent representation and the designer's knowledge. On the other hand, whereas the VaDE model learns five-dimensional latent variables, and only three of them can be interpreted, five categories can be classified by utilizing them, as shown in Fig. 9. As these latent variables are learned for the reconstruction of alternatives instead of for categorization, not all latent factors need to be interpreted for classification.

The interpretability of a classification model is an important factor in arranging the representative types and features as the products of classification. Classification trees [46] have high interpretability owing to their visualization capabilities. The method using the Gini index [47] is commonly used for learning classification trees. However, this method does not always divide alternatives into the same category under the same conditions as other alternatives. Because the relationship between the attributes and behavior within a category is not consistent in classification, designers cannot select a single category that consistently satisfies the design requirements. To address this issue, the generated alternatives are classified by binary linear classifiers while following the clustering results in this experimental application, as shown in Fig. 9. On the other hand, because a binary classifier is trained in all combinations of dividing the alternatives into two subset, the computational cost increases when the number of clusters increases. The early phase of the classification shown in Fig. 9 corresponds to the classification of clusters in the 3-cluster case shown in Fig. 7. This result indicates that the hierarchical structure of alternatives in categorization corresponds to that of classification. The computational cost can be reduced by detailing the level of the viewing aspect in deep concept identification step by step.

The extension to three-dimensional alternatives is another advanced challenge in deep concept identification. The attributes of such alternatives are represented in higher dimensions compared to two-dimensional ones. To embed such alternatives into a latent space by encoder and decoder effectively, a specific architecture is introduced according to the shape format, such as voxels [48], triangular meshes [49], and point clouds [50]. These architectures are expected to be utilized to effectively categorize three-dimensional alternatives for deep concept identification.

Design exploration is the primary objective of generative design as a whole. Although design exploration is expected to find something like extrapolation beyond the ordinary scope, generative design mostly corresponds to something like interpolation due to the nature of topology optimization. The identified concepts under the proposed framework can be utilized for further exploration when employing the idea of void mechanism [51]. In cases where entities are categorized as a product of classification for concept identification, a case in which a given category has no entity must exist. This category corresponds to void and can trigger the exploration of novel alternatives by leveraging concepts. In this implementation, the identified concepts and their relationships are arranged as decision trees. Although designers can find a void by rearranging concepts based on a decision tree, it doesn't show the void directly. For further exploration, another arrangement for finding a void should be developed. This challenge must be a potentially important future work.

## 6. Conclusion

This study proposed a concept identification framework for generative design using DL techniques. In the proposed framework, the generated alternatives were first embedded in a latent space and thereafter categorized through representation learning and clustering. Design concepts were identified as the product of classifications by interpreting the results of the categorization. The proposed framework was implemented by VaDE and logistic regression and demonstrated in a simplified conceptual design problem of a bridge structure. Although designers are required to determine the viewing aspect level by setting the number of identified concepts, this implementation presents the identified concepts and their relationships in the form of a decision tree based on a specified level.

Generative design is expected to be expanded to various design problems including fluid-based problems and heat transfer problems following the expansion of topology optimization. Although the current implementation depends on a simplified design problem of bridge structure, the proposed framework can also be applied to other design problems because of ML and DL techniques. However, the way of categorization in implementation determines the capabilities of the deep concept identification. Because VaDE categorizes data based on their overall features, the current implementation is ineffective in design problems where the design domain contains multiple areas of interest to evaluate material distribution. Extensions to these design problems are expected to be future works.

## CRediT authorship contribution statement

Ryo Tsumoto: Writing – original draft, Visualization, Validation, Software, Project administration, Methodology, Funding acquisition, Data curation, Conceptualization.

Kentaro Yaji: Writing – review & editing, Methodology, Conceptualization.

Yutaka Nomaguchi: Writing – review & editing, Methodology, Conceptualization.

Kikuo Fujita: Writing – review & editing, Supervision, Project administration, Methodology, Funding acquisition,

Conceptualization.

**Declaration of interests**

The authors declare that they have no known competing financial interests or personal relationships that could have appeared to influence the work reported in this paper.

**Data availability**

Data will be made available on request.

**Acknowledgements**

This work was supported by JST SPRING (Grant Number JPMJSP2138) and JSPS KAKENHI (Grant Number 23K28370). We would like to thank Editage (www.editage.jp) for English language editing.

**References**

[1] K. Shea, R. Aish, M. Gourtovaia, Towards integrated performance-driven generative design tools, Autom Constr 14 (2005) 253–264. https://doi.org/10.1016/j.autcon.2004.07.002.

[2] S. Krish, A practical generative design method, Computer-Aided Design 43 (2011) 88–100. https://doi.org/10.1016/j.cad.2010.09.009.

[3] J.I. Saadi, M.C. Yang, Generative design: Reframing the role of the designer in early-stage design process, Journal of Mechanical Design 145 (2023). https://doi.org/10.1115/1.4056799.

[4] S. Khan, M.J. Awan, A generative design technique for exploring shape variations, Advanced Engineering Informatics 38 (2018) 712–724. https://doi.org/10.1016/j.aei.2018.10.005.

[5] L. Zhang, Z. Li, Y. Zheng, An interactive generative design technology for appearance diversity – Taking mouse design as an example, Advanced Engineering Informatics 59 (2024) 102263. https://doi.org/10.1016/j.aei.2023.102263.

[6] J. Matejka, M. Glueck, E. Bradner, A. Hashemi, T. Grossman, G. Fitzmaurice, Dream lens: Exploration and visualization of large-scale generative design datasets, in: Proceedings of the 2018 CHI Conference on Human Factors in Computing Systems, ACM, 2018. https://doi.org/10.1145/3173574.3173943.

[7] S. Oh, Y. Jung, S. Kim, I. Lee, N. Kang, Deep generative design: Integration of topology optimization and generative models, Journal of Mechanical Design 141 (2019). https://doi.org/10.1115/1.4044229.

[8] M.P. Bendsøe, O. Sigmund, Topology Optimization, Springer Berlin Heidelberg, 2004. https://doi.org/10.1007/978-3-662-05086-6.

[9] J. Alexandersen, C.S. Andreasen, A review of topology optimisation for fluid-based problems, Fluids 5 (2020) 29. https://doi.org/10.3390/fluids5010029.

[10] A. Fawaz, Y. Hua, S. Le Corre, Y. Fan, L. Luo, Topology optimization of heat exchangers: A review, Energy 252 (2022) 124053. https://doi.org/10.1016/j.energy.2022.124053.

[11] S. Yoo, S. Lee, S. Kim, K.H. Hwang, J.H. Park, N. Kang, Integrating deep learning into CAD/CAE system: generative design and evaluation of 3D conceptual wheel, Structural and Multidisciplinary Optimization 64 (2021) 2725–2747. https://doi.org/10.1007/s00158-021-02953-9.

[12] H. Sun, L. Ma, Generative design by using exploration approaches of reinforcement learning in density-based structural topology optimization, Designs (Basel) 4 (2020) 10. https://doi.org/10.3390/designs4020010.

[13] N.Ath. Kallioras, N.D. Lagaros, MLGen: Generative design framework based on machine learning and topology optimization, Applied Sciences 11 (2021) 12044. https://doi.org/10.3390/app112412044.

[14] S. Jang, S. Yoo, N. Kang, Generative design by reinforcement learning: Enhancing the diversity of topology optimization designs, Computer-Aided Design 146 (2022) 103225. https://doi.org/10.1016/j.cad.2022.103225.

[15] Z. Wang, S. Melkote, D.W. Rosen, Generative design by embedding topology optimization into conditional generative adversarial network, Journal of Mechanical Design 145 (2023). https://doi.org/10.1115/1.4062980.

[16] G. Pahl, W. Beitz, J. Feldhusen, K.-H. Grote, Engineering Design, Springer London, 2007. https://doi.org/10.1007/978-1-84628-319-2.

[17] H. Ma, X. Chu, D. Xue, D. Chen, A systematic decision making approach for product conceptual design based on fuzzy morphological matrix, Expert Syst Appl 81 (2017) 444–456. https://doi.org/10.1016/j.eswa.2017.03.074.

[18] H.R. Fazeli, Q. Peng, Generation and evaluation of product concepts by integrating extended axiomatic design, quality function deployment and design structure matrix, Advanced Engineering Informatics 54 (2022) 101716. https://doi.org/10.1016/j.aei.2022.101716.

[19] F. Lanfermann, S. Schmitt, Concept identification for complex engineering datasets, Advanced Engineering Informatics 53 (2022) 101704. https://doi.org/10.1016/j.aei.2022.101704.

[20] H. Erhan, I.Y. Wang, N. Shireen, Harnessing design space: A similarity-based exploration method for generative design, International Journal of Architectural Computing 13 (2015) 217–236. https://doi.org/10.1260/1478-0771.13.2.217.

[21] M. Botyarov, E.E. Miller, Partitioning around medoids as a systematic approach to generative design solution space reduction, Results in Engineering 15 (2022) 100544. https://doi.org/10.1016/j.rineng.2022.100544.

[22] H.-P. Kriegel, P. Kröger, A. Zimek, Clustering high-dimensional data: A survey on subspace clustering, pattern-based clustering, and correlation clustering, ACM Trans Knowl Discov Data 3 (2009) 1–58. https://doi.org/10.1145/1497577.1497578.

[23] E. Rodrigues, D. Sousa-Rodrigues, M. de Sampayo, A.R. Gaspar, Á. Gomes, C. Henggeler Antunes, Clustering of architectural floor plans: A comparison of shape representations, Autom Constr 80 (2017) 48–65. https://doi.org/10.1016/j.autcon.2017.03.017.

[24] S. Yousif, W. Yan, Application and evaluation of a K-Medoids-based shape clustering method for an articulated design space, J Comput Des Eng 8 (2021) 935–948. https://doi.org/10.1093/jcde/qwab024.

[25] I. Goodfellow, Y. Bengio, A. Courville, Deep Learning, MIT Press, 2016.

[26] Z. Jiang, Y. Zheng, H. Tan, B. Tang, H. Zhou, Variational deep embedding: An unsupervised and generative approach to clustering, in: Proceedings of the Twenty-Sixth International Joint Conference on Artificial Intelligence, International Joint Conferences on Artificial Intelligence Organization, 2017: pp. 1965–1972. https://doi.org/10.24963/ijcai.2017/273.

[27] H.-T. Zheng, C. Borchert, Y. Jiang, A knowledge-driven approach to biomedical document conceptualization, Artif Intell Med 49 (2010) 67–78. https://doi.org/10.1016/j.artmed.2010.02.005.

[28] M. Aman, S.J. Abdulkadir, I.A. Aziz, H. Alhussian, I. Ullah, KP-Rank: a semantic-based unsupervised approach for keyphrase extraction from text data, Multimed Tools Appl 80 (2021) 12469–12506. https://doi.org/10.1007/s11042-020-10215-x.

[29] K. Minowa, K. Fujita, Y. Nomaguchi, S. Yamasaki, K. Yaji, Variational deep embedding mines concepts from comprehensive optimal designs, in: Design Computing and Cognition'20, Springer International Publishing, 2022: pp. 643–654. https://doi.org/10.1007/978-3-030-90625-2_38.

[30] K. Fujita, K. Minowa, Y. Nomaguchi, S. Yamasaki, K. Yaji, Design concept generation with variational deep embedding over comprehensive optimization, in: Volume 3B: 47th Design Automation Conference (DAC), American Society of Mechanical Engineers, 2021. https://doi.org/10.1115/detc2021-69544.

[31] L. Graening, B. Sendhoff, Shape mining: A holistic data mining approach for engineering design, Advanced Engineering Informatics 28 (2014) 166–185. https://doi.org/10.1016/j.aei.2014.03.002.

[32] R. Tsumoto, K. Fujita, Y. Nomaguchi, S. Yamasaki, K. Yaji, Classification-directed conceptual structure design based on topology optimization, deep clustering, and logistic regression, in: Volume 3A: 48th Design Automation Conference (DAC), American Society of Mechanical Engineers, 2022. https://doi.org/10.1115/detc2022-88548.

[33] H. Yoshikawa, General design theory and a CAD system, Man-Machine Communication in CAD/CAM (1981) 166–185.

[34] D. Xu, Y. Tian, A comprehensive survey of clustering algorithms, Annals of Data Science 2 (2015) 165–193. https://doi.org/10.1007/s40745-015-0040-1.

[35] S. Dreiseitl, L. Ohno-Machado, Logistic regression and artificial neural network classification models: a methodology review, J Biomed Inform 35 (2002) 352–359. https://doi.org/10.1016/s1532-0464(03)00034-0.

[36] C. Bouveyron, S. Girard, C. Schmid, High-dimensional data clustering, Computational Statistics & Data Analysis 52 (2007) 502–519. https://doi.org/10.1016/j.csda.2007.02.009.

[37] Y. Bengio, A. Courville, P. Vincent, Representation learning: A review and new perspectives, IEEE Trans Pattern Anal Mach Intell 35 (2013) 1798–1828. https://doi.org/10.1109/tpami.2013.50.

[38] D.P. Kingma, M. Welling, Auto-encoding variational bayes, in: Y. Bengio, Y. LeCun (Eds.), 2nd International Conference on Learning Representations, ICLR 2014, Banff, AB, Canada, April 14-16, 2014, Conference Track Proceedings, 2014. http://arxiv.org/abs/1312.6114.

[39] I. Higgins, L. Matthey, A. Pal, C.P. Burgess, X. Glorot, M.M. Botvinick, S. Mohamed, A. Lerchner, beta-VAE: Learning basic visual concepts with a constrained variational framework, in: 5th International Conference on Learning Representations, ICLR 2017, Toulon, France, April 24-26, 2017, Conference Track Proceedings, OpenReview.net, 2017. https://openreview.net/forum?id=Sy2fzU9gl.

[40] E. Dupont, Learning disentangled joint continuous and discrete representations, in: Proceedings of the 32nd International Conference on Neural Information Processing Systems, Curran Associates Inc., Red Hook, NY, USA, 2018: pp. 708–718.

[41] P. Linardatos, V. Papastefanopoulos, S. Kotsiantis, Explainable AI: A review of machine learning interpretability methods, Entropy 23 (2020) 18. https://doi.org/10.3390/e23010018.

[42] D.P. Billington, The Tower and the Bridge: The New Art of Structural Engineering, Basic Books, 1983.

[43] E. Andreassen, A. Clausen, M. Schevenels, B.S. Lazarov, O. Sigmund, Efficient topology optimization in MATLAB using 88 lines of code, Structural and Multidisciplinary Optimization 43 (2010) 1–16. https://doi.org/10.1007/s00158-010-0594-7.

[44] N. Otsu, A Threshold selection method from gray-level histograms, IEEE Trans Syst Man Cybern 9 (1979) 62–66. https://doi.org/10.1109/tsmc.1979.4310076.

[45] C.M. Bishop, Pattern Recognition and Machine Learning, Springer New York, NY, 2006.

[46] W. Loh, Classification and regression trees, WIREs Data Mining and Knowledge Discovery 1 (2011) 14–23. https://doi.org/10.1002/widm.8.

[47] L. Breiman, J.H. Friedman, R.A. Olshen, C.J. Stone, Classification and Regression Trees, Routledge, 2017. https://doi.org/10.1201/9781315139470.

[48] L. Xu, K. Naghavi Khanghah, H. Xu, Designing mixed-category stochastic microstructures by deep generative model-based and curvature functional-based methods, Journal of Mechanical Design 146 (2023). https://doi.org/10.1115/1.4063824.

[49] X. Li, C. Xie, Z. Sha, A predictive and generative design approach for three-dimensional mesh shapes using target-embedding variational autoencoder, Journal of Mechanical Design 144 (2022). https://doi.org/10.1115/1.4054906.

[50] N. Dommaraju, M. Bujny, S. Menzel, M. Olhofer, F. Duddeck, Evaluation of geometric similarity metrics for structural clusters generated using topology optimization, Applied Intelligence 53 (2022) 904–929. https://doi.org/10.1007/s10489-022-03301-0.

[51] T. Tomiyama, P. Breedveld, H. Birkhofer, Teaching creative design by integrating general design theory and the Pahl and Beitz methodology, in: Volume 6: 15th Design for Manufacturing and the Lifecycle Conference; 7th Symposium on International Design and Design Education, ASMEDC, 2010: pp. 707–715. https://doi.org/10.1115/detc2010-28444.